\documentclass[runningheads]{llncs}
\usepackage[T1]{fontenc}
\usepackage{graphicx}
\usepackage{amsmath}
\usepackage{multirow}
\usepackage{booktabs}
\usepackage{soul}
\usepackage{cite}
\usepackage{bm}
\usepackage{color}
\usepackage{xurl}

\usepackage{tikz}
\usepackage{caption}
\usepackage{subcaption}
\usepackage{float}
\usepackage{siunitx}
\def\checkmark{\tikz\fill[scale=0.3](0,.35) -- (.25,0) -- (1,.7) -- (.25,.15) -- cycle;} 
\begin{document}
\title{Deep Multimodal Guidance for \\Medical Image Classification}
\author{Mayur Mallya\orcidID{0000-0002-9432-4262} \and Ghassan Hamarneh\orcidID{0000-0001-5040-7448}}
\authorrunning{M. Mallya and G. Hamarneh}
\institute{Simon Fraser University, Burnaby, Canada\\
\email{\{mmallya,hamarneh\}@sfu.ca}}
\maketitle              
\begin{abstract}

Medical imaging is a cornerstone of therapy and diagnosis in modern medicine. 
However, the choice of imaging modality for a particular theranostic task typically involves trade-offs between the feasibility of using a particular modality (e.g., short wait times, low cost, fast acquisition, reduced radiation/invasiveness) and the expected performance on a clinical task (e.g., diagnostic accuracy, efficacy of treatment planning and guidance). 
In this work, we aim to apply the knowledge learned from the less feasible but better-performing (\textit{superior}) modality to guide the utilization of the more-feasible yet under-performing (\textit{inferior}) modality and steer it towards improved performance. 
We focus on the application of deep learning for image-based diagnosis.
We develop a light-weight guidance model that leverages the latent representation learned from the superior modality, when training a model that consumes only the inferior modality. We examine the advantages of our method in the context of two clinical applications: 
multi-task skin lesion classification from clinical and dermoscopic images and brain tumor classification from multi-sequence magnetic resonance imaging (MRI) and histopathology images. For both these scenarios we show a boost in diagnostic performance of the inferior modality without requiring the superior modality. Furthermore, in the case of brain tumor classification, our method outperforms the model trained on the superior modality while producing comparable results to the model that uses both modalities during inference. We make our code and trained models available at: \url{https://github.com/mayurmallya/DeepGuide}.

\end{abstract}

\section{Introduction}

Multimodal machine learning aims at analyzing the heterogeneous data in the same way animals perceive the world -- by a holistic understanding of the information gathered from all the sensory inputs. The complementary and the supplementary nature of this multi-input data helps in better navigating the surroundings than a single sensory signal. The ubiquity of digital sensors coupled with the powerful feature abstraction abilities of deep learning (DL) models has aided the massive interest in this field in the recent years~\cite{survey1, survey2, survey3, survey4, survey5}.

In clinical settings, for clinicians to make informed disease diagnosis and management decisions, a comprehensive assessment of the patient's health would ideally involve the acquisition of complementary biomedical data across multiple different modalities. For instance, the simultaneous acquisition of functional and anatomical imaging data is a common practice in the modern clinical setting~\cite{aq1, aq2}, as the former provides quantitative metabolic information and the latter provides the anatomical spatial context. Similarly, cancer diagnosis and prognosis are increasingly a result of a thorough examination of both genotypic and phenotypic modalities~\cite{pg1, pg2}.

Although the complimentary use of multiple modalities can improve the clinical diagnosis, the acquisition of modalities with stronger performance for a given task, such as diagnosis, may be less feasible due to longer wait times, higher cost of scan, slower acquisition, higher radiation exposure, and/or invasiveness. Hence, generally, a compromise must be struck. For example, 
higher anatomical or functional resolution imaging may only be possible with a modality that involves invasive surgical procedures or ionizing radiations. Other examples of this trade-off include: anatomical detail provided by computer tomography (CT) versus associated risks of cancer from repeated x-ray exposure~\cite{ct}; and rich cellular information provided by histology versus expensive, time-consuming invasive biopsy procedure with associated risks of bleeding and infections~\cite{biopsy}. Unsurprisingly, in most cases, it is the expensive modality that provides the critical piece of information for diagnosis.

For simplicity, hereinafter, we refer to the over-performing modality with less-feasible acquisition as the \textit{superior} modality, and the more-feasible but under-performing one as the \textit{inferior}. We note that a particular modality may be regarded as inferior in one context and superior in another. For example, magnetic resonance imaging (MRI) is superior to ultrasound for  delineating cancerous lesions but inferior to histopathology in deciding cancer grade.

Consequently, it would be advantageous to leverage the inferior modalities in order to alleviate the need for the superior one. However, 
this is reasonable only when the former can be as informative as the latter. To this end, we propose a novel deep multimodal, student-teacher learning-based framework that leverages existing datasets of paired inferior and superior modalities during the training phase to enhance the diagnosis performance achievable by the inferior modality during the test phase. Our experiments on two disparate multimodal datasets across several classification tasks demonstrate the validity and utility of the proposed method.

\section{Related Work}
Three sub-fields in particular relate to our work: (i) \underline{Multimodal classification:} Most of the DL based works on multimodal prediction on paired medical images focus on the classification task that involves the presence of multiple modalities at test time~\cite{medmm}. The primary focus of research being the optimal fusion strategy that aims to answer when and how to efficiently fuse the supposedly heterogeneous and redundant features. \textit{When to fuse?} While the registered multimodal image pairs allow for an input-level data fusion~\cite{inputfusion1, inputfusion2}, a majority of works rely on feature-level fusion not only due to the dimensionality mismatch at the input but also for the flexibility of fusion it offers~\cite{pathomic, mmo, sevenpt}. Additionally, some works make use of a decision-level fusion framework that leverages the ensemble learning strategies~\cite{ensemble}. \textit{How to fuse?} The most popular fusion strategy to date is the straightforward concatenation of the extracted features~\cite{concat1, sevenpt}. However, recent works aim to learn the interactions across multimodal features using strategies like the Kronecker product to model pairwise feature interactions~\cite{pathomic} and orthogonalization loss to reduce the redundancies across multimodal features~\cite{mmo}. (ii) \underline{Image Translation:} One may consider learning image-to-image translation (or style transfer) models to convert the inferior modality to the superior. However, although great success was witnessed in this field~\cite{imagetranslation1, imagetranslation2}, in the context of multimodal medical imaging, translation is complicated or non-ideal due to the difference in dimensionality (e.g. 2D to 3D) and size (e.g. millions to billions of voxels) between source and target. Additionally, the image translation only optimizes the intermediate task of translation as opposed to the proposed method that also addresses the final classification task.
(iii)  \underline{Student-Teacher (S-T) learning:}
Also referred to as knowledge distillation (KD), S-T learning aims to transfer the knowledge learned from one model to another, mostly aimed at applications with sparse or no labels, and model compression~\cite{st}. Cross-modal distillation, however, aims to leverage the modality specific representation of the teacher to distill the knowledge onto the student model. While most of such applications focus on KD across synchronized visual and audio data~\cite{audiovideo1, audiovideo2, audiovideo3}, KD methods for cross-modal medical image analysis mainly focus on segmentation~\cite{medst1, medst2, medst3}. Recently, Sonsbeek \textit{et al.}~\cite{medst_classification} proposed a multimodal KD framework for classifying chest x-ray images with the language-based electronic health records as the teacher and X-Ray images as the student network. However, unlike the proposed method, the student network in the prior works only mimics the teacher network, without explicitly incorporating its own learnt classification-specific latent representation.

To summarize, while prior works use multimodal medical images as input during inference to improve the performance, our contribution is that we leverage multimodal data during training in order to enhance inference performance with only unimodal input.

\section{Method}

\textbf{Problem formulation:} Given a training dataset $\mathcal X$ of $N$ paired images from inferior and superior modalities with corresponding ground truth target labels $\mathcal Y$, our goal is to learn a function $F$ that maps novel examples of the inferior type to target labels. Specifically, $\mathcal X = \{\mathcal X_I ,  \mathcal X_S \}$, with  $\mathcal X_I = \{ x_\mathcal{I}^i \}_{i=1}^N$ and $\mathcal X_S = \{ x_\mathcal{S}^i \}_{i=1}^N$, is the set of paired inferior and superior images, i.e. 
$(x_{\mathcal I}^i, x_{\mathcal S}^i)$ is the $i$th pair. The set of training labels is $\mathcal Y=\{y_i\}_{i=1}^N$, where $y_i \in \mathcal L$ and $\mathcal L = \{l_1, l_2, ..., l_K\}$ is the label space representing the set of all $K$ possible class labels (e.g., disease diagnoses). We represent $F$ using a deep model with parameters $\theta$, i.e.,  $\hat y=F(x_\mathcal{I};\theta)$, where $\hat y$ is the model prediction.

\begin{figure*}[!t]
\centerline{\includegraphics[width=1.00\textwidth]{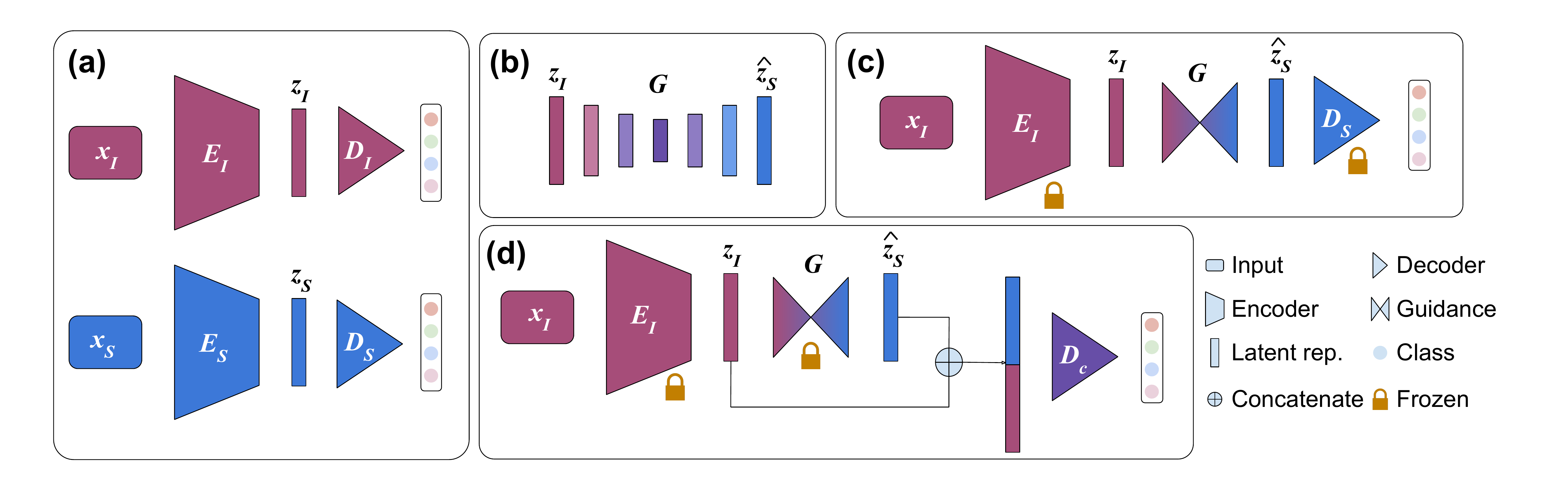}}

\caption{Overview of the multimodal guidance approach. (a) Two independent modality-specific (inferior $\mathcal I$ vs superior $\mathcal S$) classifiers are trained, each with encoder $E$---producing latent representation $z$---and decoder $D$. (b) The architecture of the guidance model $G$. (c) $G$ connects the output of the (frozen) inferior modality encoder $E_\mathcal{I}$ to the input of the (frozen) superior modality decoder $D_\mathcal{S}$. Then $G$ is trained to infer the latent representation of the superior modality from the inferior one. (d) The final model, whose input is the inferior modality alone, uses both the inferior and the estimated superior modality representations to make the final prediction via the trained combined decoder $D_c$.}
\label{fig1}
\end{figure*}

\textbf{Model optimization:} The proposed method comprises of 3 steps:
(i) Train classifiers $C_{\mathcal I}$ and $C_{\mathcal S}$ that each predicts the target label $y$ from $x_{\mathcal I}$ and from $x_{\mathcal S}$, respectively; (ii) train a guidance model $G$ to map the latent representation in $C_{\mathcal I}$ to 
that of $C_{\mathcal S}$;
(iii) construct $F$ that, first, maps $x_{\mathcal I}$ to the latent representation of $C_{\mathcal I}$, then maps that representation using $G$ to estimate the latent representation of $C_{\mathcal S}$ to perform the classification. We now describe these steps in detail.\\
\\

\noindent (i) \underline{Classifiers $C_{\mathcal I}$ and $C_{\mathcal S}$ (Figure \ref{fig1}(a)):} Given image pairs $(x_{\mathcal I}^i, x_{\mathcal S}^i)$ and ground-truth labels $y^i$, we train two independent classification models $C_{\mathcal I}$ and $C_{\mathcal S}$ on the same task. Classifier $C_{\mathcal I}$ is trained to classify images of the inferior modality $x_{\mathcal I}$, whereas $C_{\mathcal S}$ is trained to classify images of the superior modality $x_{\mathcal S}$. Denoting the predictions made by the two networks as $\hat y_{\mathcal I}^i$ and $\hat y_{\mathcal S}^i$, we have:

\begin{equation} 
\hat y_{\mathcal I}^i={C_{\mathcal I}}(x_{\mathcal I}^i) \qquad \hat y_{\mathcal S}^i={C_{\mathcal S}}(x_{\mathcal S}^i),
\label{eq1} 
\end{equation} 
\noindent where ${C_{\mathcal I}}$, and similarly ${C_{\mathcal S}}$, comprise an encoder $E$, which encodes the high-dimensional input image into a compact low-dimensional latent representation, and a decoder $D$, which decodes the latent representation by mapping it to one of the labels in  $\mathcal L$. Denoting the encoder and decoder in $C_{\mathcal I}$ as $E_{\mathcal I}$ and $D_{\mathcal I}$, respectively, and similarly $E_{\mathcal S}$ and $D_{\mathcal S}$ in $C_{\mathcal S}$, we obtain:

\begin{equation} 
\hat y_{\mathcal I}^i={D_{\mathcal I} \circ E_{\mathcal I}} (x_{\mathcal I}^i ; \theta_{E_{\mathcal I}} ) 
\qquad \hat y_{\mathcal S}^i={D_{\mathcal S} \circ E_{\mathcal S}}(x_{\mathcal S}^i ; \theta_{E_{\mathcal S}}),
\label{eq2} 
\end{equation} 

\noindent where $\circ$ denotes function composition, and
$\theta_{E_{\mathcal I}}$ and 
$\theta_{E_{\mathcal S}}$ are the encoder model parameters. The encoders produce the latent representations $z$, i.e.:

\begin{equation}
z_{\mathcal I}^i = E_{\mathcal I}(x_{\mathcal I}^i ; \theta_{E_{\mathcal I}}) 
\qquad
z_{\mathcal S}^i = E_{\mathcal S}(x_{\mathcal S}^i ; \theta_{E_{\mathcal S}}).
\end{equation}

Latent codes \noindent $z_{\mathcal I}^i$ and $z_{\mathcal S}^i$ are inputs to corresponding decoders $D_{\mathcal I}$ and $D_{\mathcal S}$. Finally, $D_{\mathcal I}(z_{\mathcal I}^i)$ and $D_{\mathcal S}(z_{\mathcal S}^i)$ yield predictions $\hat y_{\mathcal I}^i$ and $\hat y_{\mathcal S}^i$, respectively.\\
\\

\noindent (ii) \underline{Guidance Model $G$ (Figure \ref{fig1}(b-c)):} $G$ is trained to map $z_{\mathcal I}^i$ of the inferior image to $z_{\mathcal S}^i$ of the paired superior image.
Denoting the estimated latent code as $\hat z_{\mathcal S}^{i}$ and the parameters of $G$ as $\theta_{G}$, we obtain: 
\begin{equation} 
\hat z_{\mathcal S}^{i}={G}(z_{\mathcal I}^i;\theta_{G}).\label{eq:G} 
\end{equation} 

\noindent (iii) \underline {Guided Model $F$ (Figure \ref{fig1}(d)):} We incorporate the trained guidance model $G$ into the classification model of the inferior modality $C_{\mathcal I}$ so that it is steered to inherit the knowledge captured by the classifier $C_{\mathcal S}$, which was trained on the superior modality but without $C_{\mathcal I}$ being exposed to the superior image modality.
The model $F$ is thus able, during inference time, to make a prediction based solely on the inferior modality while being steered to generate internal representations that mimic those produced by models trained on superior data.  The superior modality encoder can thus be viewed as a teacher distilling its knowledge (the learned latent representation) to benefit the inferior modality student classifier.
At inference time, we need not rely only on the guided representations 
$\hat z_{\mathcal S}^{i}$ to make predictions, but rather benefit from the learned representations $z_{\mathcal I}$ from the inferior modality as well. Therefore, the two concatenated representations are used
to train a common classification decoder $D_c( [\hat z_{\mathcal S}^{i}{^\frown} z_{\mathcal I}^i] ;\theta_{D_c})$, with parameters $\theta_{D_c}$, where $^\frown$ is the concatenation operator. Thus, our final prediction $\hat y=F(x_\mathcal{I};\theta)$ is  written as follows, with
$\theta = \{\theta_{E_{\mathcal I}} , \theta_{G} , \theta_{D_c}\}$:

\begin{equation} 
\hat y=
D_c \left(
[G\left(E_{\mathcal I}(x_{\mathcal I}^i;  \theta_{E_{\mathcal I}}); \theta_{G} \right)
 ^\frown 
E_{\mathcal I}\left(x_{\mathcal I}^i ; \theta_{E_{\mathcal I}}\right)]; \theta_{D_c}
\right).
\label{eq:finalpred}
\end{equation}

\section{Experimental setup}
\textbf {Datasets:} We evaluate our method on two multimodal imaging  applications:\\

\noindent (i) \underline{RadPath 2020}~\cite{radpath} is a public dataset that was released as part of the MICCAI 2020 Computational Precision Medicine Radiology-Pathology (CPM RadPath) Challenge for brain tumor classification. The dataset consists of 221 pairs of multi-sequence MRI and digitized histopathology images along with glioma diagnosis labels of the corresponding patients. The labels include glioblastoma ($n=133$), oligodendroglioma ($n=34$), and astrocytoma ($n=54$). The MRI sequences give rise to T1, T2, T1w, and FLAIR 3D images, each of size $240 \times 240 \times 155$. The histopathology whole slide color images (WSI) are Hematoxylin and Eosin (H\&E) stained tissue specimens scanned at $20\times$ or $40\times$ magnifications, with sizes as high as $3\times80,000\times80,000$. In this scenario, the biopsy-derived WSIs form the superior modality as it provides accurate tumor diagnosis and the MRI is the non-invasive inferior modality. We divide the dataset into 165 training, 28 validation, and 28 testing splits and make 5 of such sets to test the robustness of our methods across different splits.

\noindent (ii) \underline{Derm7pt}~\cite{sevenpt} is another public dataset consisting of paired skin lesion images of clinical and dermoscopic modalities acquired from the same patients. The dataset includes 1011 pairs of images with their respective diagnosis and 7-point criteria labels for a multi-task ($n=8$) classification setup. The 7-point criteria~\cite{checklist} comprise of pigment network (PN), blue whitish veil (BWV), vascular structures (VS), pigmentation (PIG), streaks (STR), dots and globules (DaG), and regression structures (RS), each of which contributes to the 7-point score used in inferring melanoma. Both the modalities consist of 2D images of size $3\times512\times512$. As the dermoscopic images are acquired via a dermatoscope by expert dermatologists and thus reveal more detailed sub-surface 
intra- and sub-epidermal structures, they are considered the superior modality, while the clinical images may be acquired using inexpensive and ubiquitous cameras and hence regarded as the inferior modality. We adopt the pre-defined train-validation-test splits provided along with the dataset~\cite{sevenpt} and repeat our training procedure 3 times to test robustness against different random weight initializations.

\textbf{Implementation details:}
We implement the 3D DenseNet based model used by the winners of the RadPath 2019 challenge as our baseline MRI classifier~\cite{radpath_winners} and a recent multiple instance learning-based data efficient learning model, CLAM~\cite{CLAM}, as our baseline WSI classifier. For our experiments on the Derm7pt dataset, we use the pre-trained clinical and dermoscopic models provided by Kawahara \textit{et al.}~\cite{sevenpt} as base models.
In both sets of experiments, our guidance model uses an autoencoder-like architecture with a bottleneck of 256 and 512 neurons for the RadPath and Derm7pt datasets, respectively. To train the guidance model, we use the mean squared error (MSE) loss between
$z_S$ and $\hat z_S$ (see eq.~\eqref{eq:G}), and a weighted cross-entropy loss in the final combined classification model (see eq.~\eqref{eq:finalpred}) to handle the class imbalance. We use PyTorch for all our experiments on the RadPath dataset and TensorFlow for our experiments on Derm7pt dataset, as we build on top of the Keras-based pre-trained models \cite{sevenpt}. While the base models for RadPath were trained on a multi-GPU cluster, the rest of the models were trained on an NVIDIA GeForce GTX 1080 Ti GPU. We provide the values of optimal hyperparameters used for different experiments as supplementary table, which reports the batch sizes, optimizer parameters, loss weights, and early stopping parameters.

\textbf{Evaluation metrics:}
Since our experiments involve imbalanced classes, we use balanced accuracy (BA) in conjunction with the micro F1 score. While the macro-averaging of class-wise accuracy values in BA gives equal importance to all classes, the micro-averaged F1 score represents the overall correctness of the classifier irrespective of the class performance~\cite{metrics}. Additionally, for the binary task of melanoma inference, we use the AUROC score.

\textbf{Melanoma Inference:}
Similar to Kawahara \textit{et al.}~\cite{sevenpt}, in addition to direct diagnosis of skin lesions, we infer melanoma from the 7-point criteria predictions~\cite{checklist}. Based on these predictions, we compute the 7-point score and use the commonly used thresholds of $t={1,3}$ for inferring melanoma~\cite{sevenpt,checklist}. Finally, we compute the AUROC score to compare the overall performance of the classifiers.

\section{Results and Discussion}

\begin{table}[t]

\caption{RadPath results. Radiology MRI sequences (R) are guided using pathology (P). Only the best performing model using ALL MR sequences is used, explaining the `-' in row 1. In row 2, MR is not used, only P, hence `*'. R is the inferior ($\mathcal I$) and P is the superior modality ($\mathcal S$).}
\label{radpath_results}
\setlength{\tabcolsep}{3pt}
\centering
\begin{tabular}{clcccccccccc}
\toprule

& \multirow{2}{*}{Method} & \multicolumn{2}{c}{T1} & \multicolumn{2}{c}{T2} & \multicolumn{2}{c}{T1w} & \multicolumn{2}{c}{FLAIR} & \multicolumn{2}{c}{ALL} \\ 

\cmidrule(lr){3-4} \cmidrule(lr){5-6} \cmidrule(lr){7-8} \cmidrule(lr){9-10} \cmidrule(lr){11-12} 
& & {BA$\uparrow$} & {F1$\uparrow$} & {BA$\uparrow$} & {F1$\uparrow$} & {BA$\uparrow$} & {F1$\uparrow$} & {BA$\uparrow$} & {F1$\uparrow$} & {BA$\uparrow$} & {F1$\uparrow$}\\
\cmidrule(lr){1-12}
1. & $P+R$ & - & - & - & - & - & - & - & - & 0.777 & 0.821 \\
2. & $P$~\cite{CLAM}  & * & * & * & * & * & * & * & * & 0.729 & 0.792 \\
\cmidrule(lr){1-12}
3. & $R$~\cite{radpath_winners}  & 0.506 & 0.578 & 0.657 & 0.728 & 0.641 & 0.735 & 0.534 & 0.621 & 0.729 & 0.771\\
4. & $G(R)$     & 0.443 & 0.578 & 0.559 & 0.692 & 0.564 & 0.692 & 0.419 & 0.571 & 0.650 & 0.764\\
5. & $G(R)+R$    & 0.587 & 0.585 & 0.731 & 0.757 & 0.654 & 0.742 & 0.648 & 0.635 & 0.752 & 0.799\\
\cmidrule(lr){1-12}
6. & $\Delta$ $(\%)$ & $+$16.0 & $+$1.2 & $+$11.2 & $+$3.9 & $+$2.0 & $+$0.9 & $+$21.3 & $+$2.2 & $+$3.1 & $+$3.6 \\

\bottomrule
\end{tabular}
\end{table}

The results on RadPath and Derm7pt datasets are in Tables \ref{radpath_results} and \ref{skin_results}, respectively, where we report the classification performance, BA and F1, across multiple tasks, under different input modalities and guidance strategies: when using both superior and inferior modalities as input (referred to as $\mathcal S+\mathcal I$); superior alone ($\mathcal S$); inferior alone without guidance ($\mathcal I$); guided inferior ($G(\mathcal I)$); and guided inferior with inferior ($G(\mathcal I)+\mathcal I$). For RadPath, radiology (R) is inferior and can be either of T1, T2, T1w, FLAIR, or ALL combined, and pathology (P) is superior while for Derm7pt, clinical (C) is inferior and dermoscopic (D) is superior.

\bm{$\mathcal S$} \textbf{outperforms} \bm{$\mathcal I$} (row 2 vs 3): The results from the baseline models of both experiments confirm that the superior modality is more accurate for disease diagnosis. Table \ref{radpath_results}, shows that the classifier $P$ significantly outperforms the individual MRI classifiers while being marginally better than the ALL MRI classifier. Similarly, from Table \ref{skin_results}, the classifier $D$ outperforms the classifier $C$ across all the 7-point criteria, diagnosis, and the overall melanoma inference.

\bm{$\mathcal S+\mathcal I$} \textbf{outperforms} \bm{$\mathcal S$} (row 1 vs 2): When using both modalities for classification, in case of RadPath, we observe the expected improvement over the classification using only superior, affirming the value added by MRI. However, the combined skin classifier does not concretely justify the addition of inferior (as also shown earlier~\cite{abhishek2021predicting,sevenpt}). We attribute this to the redundancy of information from the clinical images, which further adds to our motivation of guiding the inferior modality using the superior modality.

\bm{$G(\mathcal I)$} \textbf{alone does not outperform} \bm{$\mathcal I$} (row 3 vs 4): Our guided model, trained with the solitary goal of learning a mapping from the inferior features to superior using an MSE loss, performs worse than the original inferior model in both the datasets. This is not surprising as the poorer performance of this method suggests an imperfect reconstruction of the superior features, which can be attributed to the small size of the datasets. Consequently, our proposed method does not ignore the importance of the inferior modality, as shown next.

\begin{table}[t]
\caption{Derm7pt results. The 7-point criteria, diagnosis (DIAG), and melanoma (MEL) are inferred. Clinical (C) is guided by Dermoscopic (D). C is the inferior ($\mathcal I$) and D is the superior modality ($\mathcal S$).}
\label{skin_results}
\setlength{\tabcolsep}{3pt}
\centering
\resizebox{\textwidth}{!}{
\begin{tabular}{clcccccccccccc}
\toprule

& \multirow{2}{*}{Method} & & \multicolumn{7}{c}{7-point criteria} & \multirow{2}{*}{DIAG} & \multicolumn{3}{c}{MEL Inference} \\
\cmidrule(lr){4-10} \cmidrule(lr){12-14}
& & & PN & BWV & VS & PIG & STR & DaG & RS &  & $t=1$ & $t=3$ & AUROC \\ 
\cmidrule(lr){1-14}

\multirow{2}{*}{1.} & \multirow{2}{*}{$D+C$~\cite{sevenpt}} & BA$\uparrow$
 & 0.686 & 0.772 & 0.472 & 0.520 & 0.611 & 0.597 & 0.731 & 0.484 & 0.673 & 0.716 & \multirow{2}{*}{0.788} \\ 
 & & F1$\uparrow$ & 0.693 & 0.882 & 0.816 & 0.630 & 0.734 & 0.617 & 0.783 & 0.688 & 0.591 & 0.788 & \\
\cmidrule(lr){4-14}

\multirow{2}{*}{2.} & \multirow{2}{*}{$D$~\cite{sevenpt}} & BA$\uparrow$
 & 0.666 & 0.809 & 0.552 & 0.573 & 0.621 & 0.583 & 0.719 & 0.635 & 0.656 & 0.702 & \multirow{2}{*}{0.764} \\ 
 & & F1$\uparrow$ & 0.688 & 0.859 & 0.790 & 0.632 & 0.716 & 0.596 & 0.770 & 0.716 & 0.586 & 0.762 & \\

\cmidrule(lr){1-14}

\multirow{2}{*}{3.} & \multirow{2}{*}{$C$~\cite{sevenpt}} & BA$\uparrow$
 & 0.585 & 0.690 & 0.513 & 0.484 & 0.581 & 0.530 & 0.687 & 0.438 & 0.663 & 0.691 & \multirow{2}{*}{0.739} \\ 
 & & F1$\uparrow$ & 0.584 & 0.775 & 0.747 & 0.571 & 0.625 & 0.528 & 0.714 & 0.604 & 0.576 & 0.716 & \\ 

\cmidrule(lr){4-14}

\multirow{2}{*}{4.} & \multirow{2}{*}{$G(C)$} & BA$\uparrow$
 & 0.563 & 0.664 & 0.395 & 0.460 & 0.509 & 0.503 & 0.653 & 0.330 & 0.675 & 0.630 & \multirow{2}{*}{0.716}\\
 & & F1$\uparrow$ & 0.588 & 0.822 & 0.803 & 0.611 & 0.681 & 0.501 & 0.763 & 0.616 & 0.682 & 0.744 & \\

\cmidrule(lr){4-14}

\multirow{2}{*}{5.} & \multirow{2}{*}{$G(C)+C$} & BA$\uparrow$
 & 0.591 & 0.696 & 0.528 & 0.502 & 0.548 & 0.540 & 0.695 & 0.447 & 0.691 & 0.704 & \multirow{2}{*}{0.751} \\ 
& & F1$\uparrow$ & 0.613 & 0.810 & 0.749 & 0.595 & 0.630 & 0.552 & 0.744 & 0.618 & 0.629 & 0.736 &  \\ 

 \cmidrule(lr){1-14}

\multirow{2}{*}{6.} & \multirow{2}{*}{$\Delta$ ($\%$)} & BA$\uparrow$
 & $+$1.0 & $+$0.8 & $+$2.9 & $+$3.7 & ($-$5.6) & $+$1.8 & $+$1.1 & $+$2.0 & $+$4.2 & $+$1.8 & \multirow{2}{*}{$+$1.6} \\ 
 & & F1$\uparrow$ & $+$4.9 & $+$4.5 & $+$0.2 & $+$4.2 & $+$0.8 & $+$4.5 & $+$4.2 & $+$2.3 & $+$9.2 & $+$2.7 &  \\ 
\bottomrule

\end{tabular}
}
\end{table}

\bm{$G(\mathcal I)+\mathcal I$} \textbf{outperforms} \bm{$\mathcal I$} (row 3 vs 5):
Our proposed model (row 5), which retains the inferior inputs alongside the reconstructed superior features (using the guidance model), performs better than the baseline method (row 3) -- that takes inferior inputs without leveraging the superior features -- across all 5 radiology models and 7 of the 8 clinical-skin models. Row 6 in both tables reports the percentage improvement $\Delta$ in the performance achieved by the proposed method (row 5) over the baseline inferior classifier (row 3).

\bm{$G(\mathcal I)+\mathcal I$} \textbf{reaches performance of} \bm{$\mathcal S+\mathcal I$} \textbf{for RadPath} (row 1 vs 5):
Moreover, our proposed model with ALL sequences (Table \ref{radpath_results} row 5) outperforms the superior modality model (row 2) while being comparable to the model that takes both the inferior and superior modalities as inputs during the test time (row 1), essentially alleviating the need for the superior inputs during the inference. However, in the case of Derm7pt, we observe that the improvement in performance over the baseline method does not approach the performance of the superior model, thus underscoring the superiority of information provided by the dermoscopic images over clinical. 
Additionally, we hypothesize that the performance of $\mathcal S + \mathcal I$ forms an upper bound on the performance achievable by $G(\mathcal I) + I$ as the guidance ($G(\mathcal I)$) aims to mimic the latent representation of the superior modality ($\mathcal S$).

\section{Conclusion}

Motivated by the observation that for a particular clinical task the better-performing medical imaging modalities are typically less feasible to acquire, in this work we proposed a student-teacher method that distills knowledge learned from a better-performing (superior) modality to guide a more-feasible yet under-performing (inferior) modality and steer it towards improved performance. Our evaluation on two multimodal medical imaging based diagnosis tasks (skin and brain cancer diagnosis) demonstrated the ability of our method to boost the classification performance when only the inferior modality is used as input. We even observed (for the brain tumour classification task) that our proposed model, using guided unimodal data, achieved results comparable to a model that uses both superior and inferior multimodal data, i.e. potentially alleviating the need for a more expensive or invasive acquisition. Our future work includes extending our method to handle cross-domain continual-learning and testing on other applications.

\subsubsection{Acknowledgements} 

We thank Weina Jin, Kumar Abhishek, and other members of the Medical Image Analysis Lab, Simon Fraser University, for the helpful discussions and feedback on the work. This project was funded by the Natural Sciences and Engineering Research Council of Canada (NSERC), BC Cancer Foundation-BrainCare BC Fund, and the Mitacs Globalink Graduate Fellowship. Computational resources were provided by Compute Canada (\url{www.computecanada.ca}).

\bibliographystyle{splncs04}
\bibliography{refs}


\newpage
\title{Supplementary: Deep Multimodal Guidance \\for Medical Image Classification}
\author{Mayur Mallya \and Ghassan Hamarneh}
\authorrunning{M. Mallya and G. Hamarneh}
\institute{Simon Fraser University, Burnaby, Canada\\
\email{\{mmallya,hamarneh\}@sfu.ca}}

\maketitle   

\begin{table}[h]
\caption{Hyperparameters used for our experiments. BS: batch size, optim: optimizer, LR: learning rate, WRS: weighted random sampler, $*$: all tasks.}
\label{hyperparameters}
\setlength{\tabcolsep}{3pt}
\centering
\resizebox{\textwidth}{!}{
\begin{tabular}{cccccccccccc}
\toprule
Experiment & task & GPU & BS & epochs & patience & network & optim & LR & loss & loss weights & WRS \\
\cmidrule(lr){1-12}

$P+R$ & - & 11G & 50 & 500 & 200 & Linear & SGD & $1\times10^{-3}$ & CE & [1.0, 1.7, 1.6] & \checkmark \\

$P$ & - & 4$\times$12G & 1 & 200 & 20 & CLAM & Adam & $1\times10^{-4}$ & CE & - & \checkmark \\

$R$ & $*$ & 32G & 10 & 600 & 50 & DenseNet121 & Adam & $2\times10^{-4}$ & CE & - & \checkmark\\

$G(R)$ & * & 11G & 50 & 150 & - & Linear & SGD & $5\times10^{-1}$ & MSE & - & - \\

$G(R)+R$ & * & 11G & 50 & 500 & 200 & Linear & SGD & $1\times10^{-3}$ & CE & [1.0, 1.7, 1.6] & \checkmark \\
\cmidrule(lr){1-12}

$D+C$ & * & \multicolumn{10}{c}{Pretrained model} \\

$D$ & * & \multicolumn{10}{c}{Pretrained model} \\

$C$ & * & \multicolumn{10}{c}{Pretrained model} \\

$G(C)$ & * & 11G & 100 & 500 & - & Linear & SGD & $5\times10^{-1}$ & MSE & - & - \\

\multirow{8}{*}{$G(C)+C$} 
& PN  & 11G & 100 & 500 & 200 & Linear & SGD & $1\times10^{-4}$ & CE & [0.9, 0.9, 1.8] & - \\
& BWV & 11G & 100 & 500 & 200 & Linear & SGD & $1\times10^{-4}$ & CE & [0.6, 12.0] & - \\
& VS  & 11G & 100 & 500 & 200 & Linear & SGD & $1\times10^{-4}$ & CE & [0.01, 20.0, 50.0] & - \\
& PIG & 11G & 100 & 500 & 200 & Linear & SGD & $1\times10^{-4}$ & CE & [0.2, 45.0, 1.1] & - \\
& STR & 11G & 100 & 500 & 200 & Linear & SGD & $1\times10^{-4}$ & CE & [0.06, 50.0, 4.0] & - \\
& DaG & 11G & 100 & 500 & 200 & Linear & SGD & $1\times10^{-4}$ & CE & [0.3, 0.9, 0.9] & - \\
& RS  & 11G & 100 & 500 & 200 & Linear & SGD & $1\times10^{-4}$ & CE & [0.25, 6.0] & - \\
& DIAG & 11G & 100 & 500 & 200 & Linear & SGD & $1\times10^{-4}$ & CE & [15.0, 0.07, 1.4, 4.0, 15.0] & - \\
\bottomrule

\end{tabular}
}
\end{table}

\begin{figure}
\centering
\begin{subfigure}{0.49\textwidth}
  \centering
  \includegraphics[width=.8\linewidth]{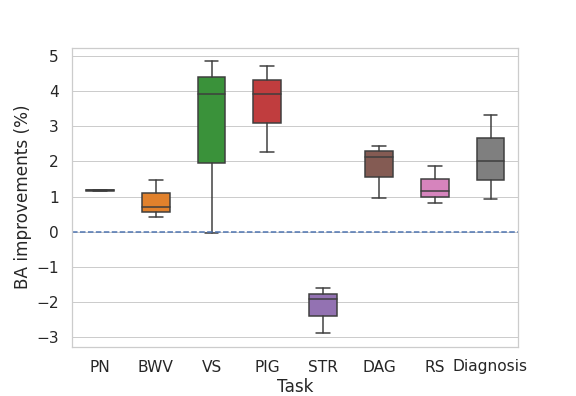}  
  \caption{$\%$ BA improvements in Derm7pt}
  \label{fig:sub-first}
\end{subfigure}
\begin{subfigure}{0.49\textwidth}
  \centering
  \includegraphics[width=.8\linewidth]{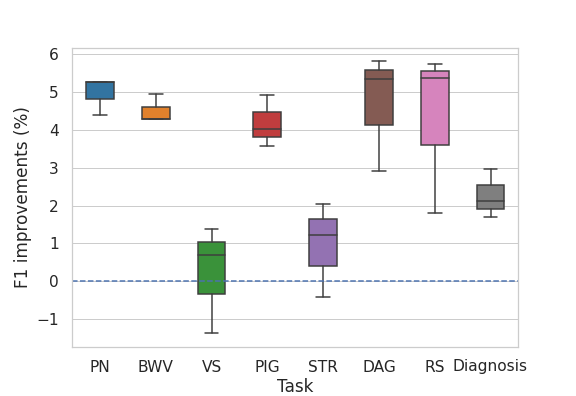}  
  \caption{$\%$ F1 improvements in Derm7pt}
  \label{fig:sub-second}
\end{subfigure}

\begin{subfigure}{0.49\textwidth}
  \centering
  \includegraphics[width=.8\linewidth]{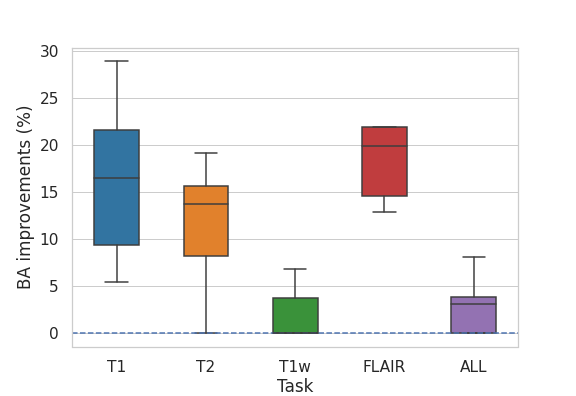}  
  \caption{$\%$ BA improvements in RadPath}
  \label{fig:sub-third}
\end{subfigure}
\begin{subfigure}{0.49\textwidth}
  \centering
  \includegraphics[width=.8\linewidth]{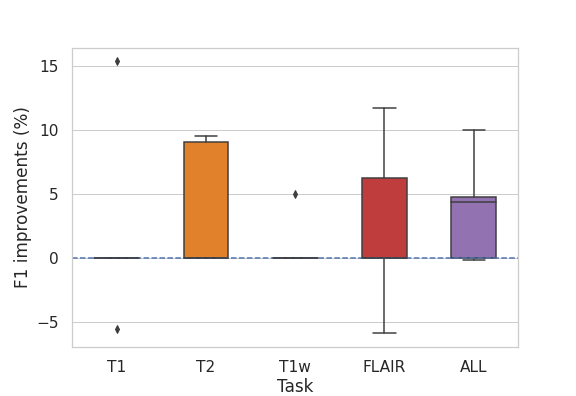}  
  \caption{$\%$ F1 improvements in RadPath}
  \label{fig:sub-fourth}
\end{subfigure}
\caption{Improvements of \bm{$G(\mathcal I)+\mathcal I$} over \bm{$\mathcal I$}. BA: balanced accuracy, F1: F1 micro.}
\label{fig:fig}
\end{figure}

\begin{table}[h]
\caption{Statistical significance of multimodal guidance for the Derm7pt dataset (size of the test set = 392). We use the McNemar's significance test to compare the guided model ($G(\mathcal I) + \mathcal I$) with the baseline clinical model ($\mathcal I$). The table shows the p-values corresponding to the McNemar's test for each of the tasks and the results are considered to be statistically significant for $p < 0.05$. }
\label{stat_sig}
\setlength{\tabcolsep}{3pt}
\centering
\begin{tabular}{cccc}
\toprule
PN & BWV & VS & PIG  \\
\num{1.55e-1} & \num{2.94e-2} & \num{5.71e-1} & \num{1.17e-1} \\
\cmidrule(lr){1-4}
STR & DaG & RS & DIAG\\
\num{6.43e-1} & \num{1.55e-1} & \num{1.95e-2} &  \num{4.57e-1} \\
\bottomrule
\end{tabular}
\end{table}

\end{document}